\documentclass[letterpaper, 10 pt, conference]{ieeeconf} 
\usepackage[utf8]{inputenc}
\usepackage{graphicx}
\usepackage{hyperref}
\usepackage{comment}
\usepackage[text=gray,background=white]{callouts}
\usepackage{tikz}
\usepackage{float}
\usepackage{amsmath}
\IEEEoverridecommandlockouts                              
\title{\LARGE \bf
A Novel Approach to Tomato Harvesting Using a Hybrid Gripper with Semantic Segmentation and Keypoint Detection
}

\author{Shahid Ansari$^{1}$, Mahendra Kumar Gohil$^{1}$ and Bishakh Bhattacharya$^{1}$
\thanks{*This work was supported by Department of Science and Technology, India}
\thanks{$^{1}$Shahid is a research scholar with the Department of Mechanical Engineering, Indian Institute of Technology Kanpur, 208016 Kanpur,India
        {\tt\small shahid@iitk.ac.in}}%
\thanks{$^{1}$Mahendra is a research scholar with the Department of Mechanical Engineering, Indian Institute of Technology Kanpur, 208016 Kanpur,India
        {\tt\small mgohil@iitk.ac.in}}%
\thanks{$^{1}$Bishakh Bhattacharya is with the Department of Mechanical Engineering, Indian Institute of Technology Kanpur, 208016 Kanpur, India
        {\tt\small bishakh@iitk.ac.in}}%
}

\begin{document}
\maketitle
\thispagestyle{empty}
\pagestyle{empty}

\begin{abstract}
Current agriculture and farming industries are able to reap advancements in robotics and automation technology to harvest fruits and vegetables using robots with adaptive grasping forces based on the compliance or softness of the fruit or vegetable. A successful operation depends on using a gripper that can adapt to the mechanical properties of the crops. This paper proposes a new robotic harvesting approach for tomato fruit using a novel hybrid gripper
with a soft caging effect. It uses its six flexible passive auxetic structures based on fingers with rigid outer exoskeletons for good gripping strength and shape conformability. The gripper is actuated through a scotch-yoke mechanism using a servo motor. To perform tomato picking operations through a gripper, a vision system based on a depth camera and RGB camera implements the fruit identification process. It incorporates deep learning-based keypoint detection of the tomato's pedicel and body for localization in an occluded and variable ambient light environment and semantic segmentation of ripe and unripe tomatoes. In addition, robust trajectory planning of the robotic arm based on input from the vision system and control of robotic gripper movements are carried out for secure tomato handling. The tunable grasping force of the gripper would allow the robotic handling of fruits with a broad range of compliance.
\end{abstract}

\section{INTRODUCTION}
Precision agriculture and smart farming technologies are becoming more prevalent due to their potential to cater to growing demand and meet the needs of the global food supply. Smart farming technologies involve integrating technology and data-driven agriculture applications to improve crop yield and influence the high quality of food products\cite{gupta2020security}. 

Autonomous and intelligent robot systems are called for agricultural industrialization to increase harvesting precision and productivity. These systems will assist farm factories in resolving the conflict between the high cost of labor and the issues with product quality.
In recent years, conventional harvesting techniques that use combine harvesters, reapers, or trunk shakers have been developed\cite{shojaei2021intelligent,cho2014using}. However, most of these machine-like harvesting equipment are inappropriate for soft and delicate crops like tomatoes, strawberries, and apples since the high force applied by the gripper may impact and harm the harvested fruits. Hence, techniques like "Selective Harvesting," in which crops are carefully picked one by one like a human would, are best for such crops. The challenges of selective harvesting originate from the demands of locating the crop and determining its position and the pedicel to be cut. In the actual world, crops like tomatoes are densely packed around and obscured by branches and foliage, making it challenging to identify the target tomato from its surroundings\cite{chen2015reasoning}. Research on robotic harvesting technology for soft fruits and vegetables is, therefore, essential\cite{comba2010robotics,ling2019dual}. Extensive research is happening to find an effective robotic system design that can overcome numerous challenges in addition to changing farming methods to automate and intelligently harvest fruits and vegetables utilizing robots.\cite{gao2022development}proposed an end effector consisting of two clamping fingers mounted over housing and powered by a limiting block that moves up and down over it through the piston rod of a telescopic cylinder. A gear set driven by a rotating pneumatic cylinder provides rotational motion to fingers.

For conformal grasping of soft objects,\cite{tawk20223d} developed a 3D-printed modular soft pneumatic gripper incorporated with mechanical metamaterials. The gripper creates bending by pneumatically activating the soft monolithic pneumatic digits.\cite{Kaur2019toward} elucidates the methodology and implementation of a robotic finger with an architecture design utilizing 3D-printed metamaterials. The finger design incorporates compliant auxetic joints, an octet body structure for adaptive grasping, and a pressure sensor that is embedded inside .\cite {kondo2010development}proposed an end-effector that can harvest individual tomatoes and tomato clusters. This end effector comprises two parallel pairs of digits separated by a distance and actuated by a pneumatic piston-cylinder actuator. It can hold and cut peduncles by sensing the peduncle with strain sensors mounted on the lower digits. \cite{ansari2022design} proposed a novel soft gripper for tomato harvesting with 3D-printed flexible fingers that form a cage-like structure when a servo motor with thin, rigid cables pulls together all the flexible beams. In addition, it has a mechanism for separating the target tomato from other tomatoes in the cluster and a servo-driven Iris-based pedicel-cutting mechanism. \cite{liu2018soft} proposed a soft robotic gripper module designed for delicately grasping fruits. The gripper features 3D printed compliant fingers, making it underactuated and sensor-less. The gripper module, actuated by a single linear actuator, is mounted on a robot arm and integrated with machine vision for automated fruit picking and placing.
To handle fragile objects\cite{baker2023star} proposed an anthropomorphic soft robotic gripper driven by twisted string actuators (TSAs) with a monolithic structure with a 3-DOF thumb and four 2-DOF fingers, enabling more natural grasping motions. 

Semantic segmentation and keypoint detection have become crucial techniques in computer vision. In agriculture, these methodologies are highly advantageous for analyzing fruits, encompassing various tasks such as fruit detection, segmentation, and picking point localization.

 In a notable study conducted by Zhou et al. \cite{Zhou2023}, a novel approach was introduced to detect and select dragon fruits. The researchers employed the YOLOv7 algorithm to localize and classify the fruits effectively while also incorporating the PSP-Ellipse method to identify the fruits' endpoints accurately. This methodology showcases a promising advancement in fruit detection and selection. Similarly, Liang et al. \cite{Liang2022} have put forth an approach to assess defective apples' quality in real-time effectively. This method entails the utilization of a hybrid framework that combines semantic segmentation, leveraging the BiSeNet V2 deep learning network, with a pruned YOLO V4 network.

Furthermore, extensive research has been conducted to explore the practical implementation of the aforementioned techniques for distinct types of fruits. In a recent study by Tafuor et al.  \cite{10.1109/ICRA46639.2022.9812303}, a notable contribution was made in the form of two innovative strawberry datasets. These datasets were meticulously annotated to include crucial information such as picking points, key points, and the weight and size of the berries. The authors have successfully demonstrated the efficacy of keypoint detection in the context of point localization for picking and grasping tasks. Additionally, they have introduced a pioneering baseline model for weight estimation, which exhibits superior performance compared to numerous state-of-the-art deep networks.

In their seminal work, Yan et al. \cite{Yan2023} introduced a pioneering dome-type planted pumpkin autonomous harvesting framework. This framework incorporates a novel keypoint detection method specifically designed for identifying the grasping and cutting points, utilizing an instance segmentation architecture.

In this work, we have proposed a novel robotic gripper system for tomato harvesting using deep learning-based semantic segmentation and keypoint detection. The objective of this work focuses on the following key aspects:
1. A detailed explanation of the robotic gripper mechanical architecture with the input-output mathematical relationship between the motor torque provided to the gripper mechanism and the grasping force required to hold the tomato in place through the soft-auxetic structure-based fingers is established using the principle of virtual work.
2. Performance analysis of the overall system is carried out regarding the various operations during the tomato-picking process. 
3. Implement a vision system with deep learning-based Semantic segmentation of ripe and unripe tomatoes and keypoint detection of pedicel and tomato center in visible and occluded conditions for localization in a variable ambient light environment
4. Robust trajectory planning of the robotic arm based upon
the vision system input and control of robotic gripper actions for safe tomato handling in place.
\section{DESIGN}
\begin{figure*}[htbp]
       \centering
       \includegraphics[width=0.8\linewidth]{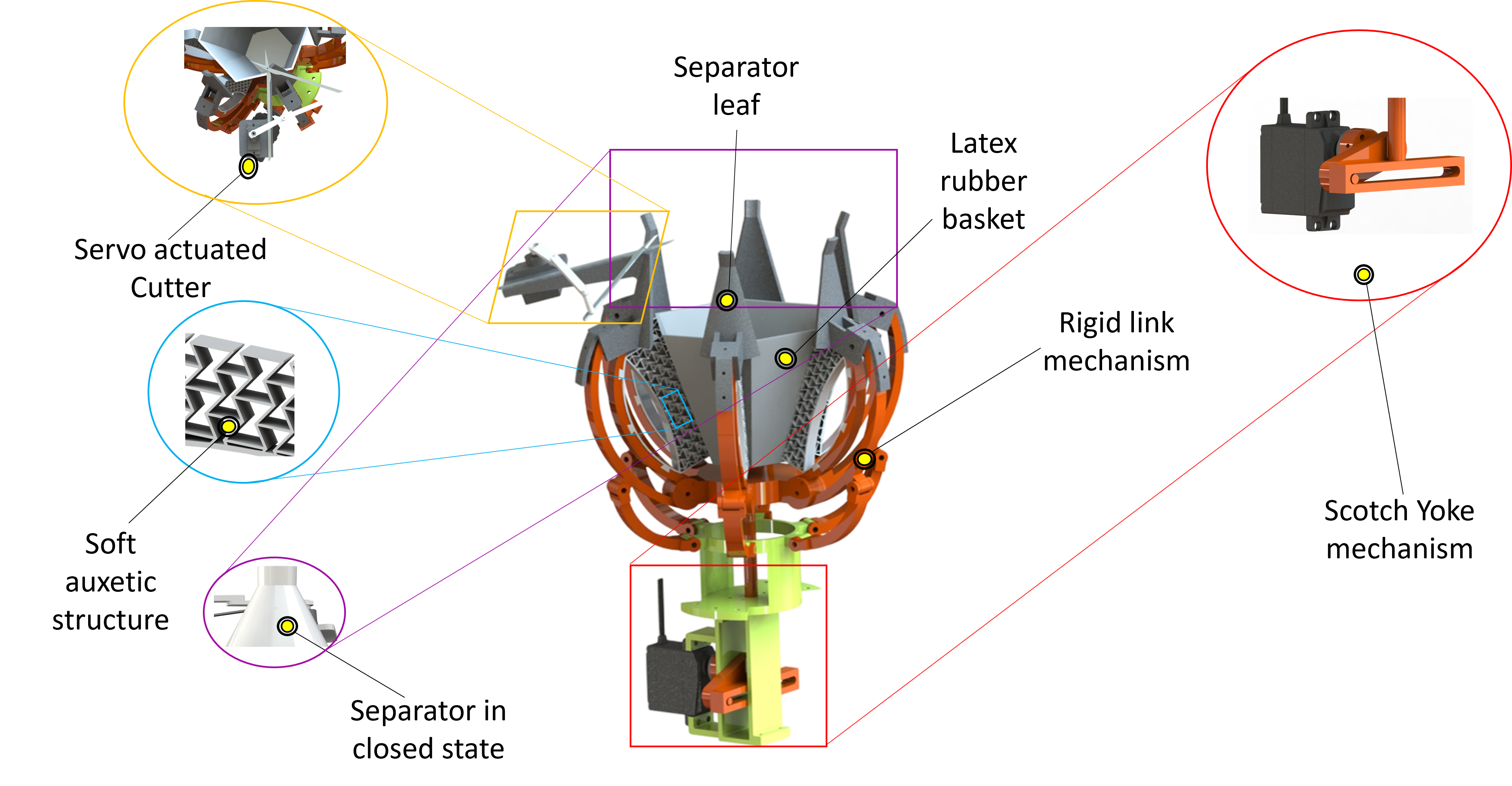}
      \caption{The figure shows a 3D CAD model of the robotic gripper mechanical design with details of the main parts of the gripper, which remains active throughout the tomato harvesting process. The top left shows the cutting mechanism, which is actuated by a micro servo motor; the middle left shows the soft auxetic structure for gentle grasp; the bottom left shows the separator leaves in a closed state forming a conical frustum for tomato separation, and top right shows the servo actuated scotch yoke mechanism to drive the gripper fingers.}
      \label{Gripperassembly_CAD}
\end{figure*}
The proposed robotic gripper incorporates the combination of a 2D re-entrant soft auxetic structure, a type of metamaterial, and the rigid linkage, which forms an exoskeleton for the inner part of the fingers for enhanced gripping strength over the object. Auxetic structures with a negative Poisson's ratio have the unique ability to expand laterally when stretched longitudinally and contract longitudinally when compressed laterally due to internal re-entrant honeycomb structure with rotating hinged beams compared to conventional materials\cite{mir2014review}. This property allows them to adapt to irregular shapes and sizes of objects, making them well-suited for grasping soft and deformable objects by avoiding the development of localized stress concentrations. The design idea of the gripper is to have a firm hold over the tomato through a gentle and conformable grasp using the caging effect of the six fingers. Here, the auxetic structure is the combination of the pattern of honeycomb re-entrant design and curved leaf spring, which have the effect of shape conformability over the object and the spring effect of having the desired stiffness. To hold the tomato in place and avoid falling to the outside of the gripper, a basket made of latex is bonded over the auxetic structures, creating a variable, encompassing 3D space for capture. The system also consists of the six separator leaves, which, when closed during actuation, make a conical frustum that, when forced towards the tomato cluster, will separate the target tomato from tomatoes in its vicinity by opening the leaves slightly. Also, the cutter is mounted over one of the separator leaves, which, when actuated with the help of a micro servo, will cut the pedicel of the tomato after the separation stage. The mechanism of the gripper, along with the soft auxetic structures, is shown in Fig.\ref{Gripperassembly_CAD}.
\subsection{Materials and components}
The outer structure of the gripper consists of rigid links, and the driving Scotch Yoke mechanism is 3D printed using PLA(Poly Lactic acid) filament on Creality ender3 V2 3D printer, while the soft auxetic structure is 3D printed with TPU(Thermoplastic Polyurethane) filament on Ultimaker 3D printer. The holding basket is custom-built using 1mm thin latex rubber to provide a small elastic resistance to the gripper while opening it. The motor used for driving the gripper mechanism is Orange OT5316M 7.4V 15kg.cm Metal Gear Digital Servo Motor While for the actuating the cutter, we used Align DS426M digital micro servo. For tomato detection and keypoint identification, the ZED2i camera by Stereolabs is used. For manipulation tasks, a five-degrees-of-freedom Viperx300 robotic arm by InterbotiX is used.
\subsection{Torque requirement calculation}
To know the amount of torque required to grasp the tomato using the gripper with soft auxetic structure-based fingers, mathematical analysis is performed using the virtual work principle, which will directly establish an input-output relationship. According to the principle of virtual work, the virtual work done by external active forces on an ideal mechanical system in equilibrium is zero for all virtual displacements consistent with the constraints (restriction of the motion by the supports). The work done by all the reactive and internal forces will be zero.\\

\begin{equation}
\begin{aligned}
   dU & = 0
\end{aligned}
\end{equation}
\begin{gather}
  T\delta\theta + P\delta x_{m} + P\delta y_{m} = 0 \\
\begin{aligned}
  x_{m} &= r\cos\theta + l_{s} + l_{DM}\cos\xi \\
  y_{m} &= l_{p} + l_{DM}\sin \xi \\
  \delta x_{m} &= -r\sin \theta \delta \theta - l_{DM}\sin \xi \delta \xi \\
  \delta y_{m} &= l_{DM}\cos \xi \delta \xi \\
\end{aligned} \\
\begin{aligned}
  r\cos\theta + f &= a + e\cos\beta + d\cos\xi \\
  c + e\sin\beta &= b + d\sin\xi \\
\end{aligned} \\
\xi = 180^o - \beta - \gamma \\
k = \sqrt{(r\cos\theta + f - a)^2 + (c - b)^2} \\
u = \tan^{-1}\left(\frac{c - b}{r\cos\theta + f - a}\right) \\
\beta = \cos^{-1}\left(\frac{1}{2ek}\left[(r\cos\theta + f - a)^2 + (c - b)^2 - d^2\right]\right) - u \\
T = Pl_{DM}\sin\xi\frac{\delta\xi}{\delta\theta} + Pr\sin\theta
\end{gather}
Here torque $T$ is in N-mm, Force $P$ is in Newton, $x_m, y_m, l_{p}, l_{DM}, r, a, b, c, d, e, f, k$ are in mm while $\alpha.\beta,\gamma,\zeta,\theta,u$ are in degree.
From these equations (1–9), the relationship between the known grasping force to hold the tomato and the torque required from the motor to hold it is established. This analysis is for the single finger; however, since six fingers are located on the gripper symmetrically, the total amount of torque required to provide the desired force will be six times the torque which is obtained from this expression.
\begin{figure}[htbp]
      \centering
      \includegraphics[width=0.7\linewidth]{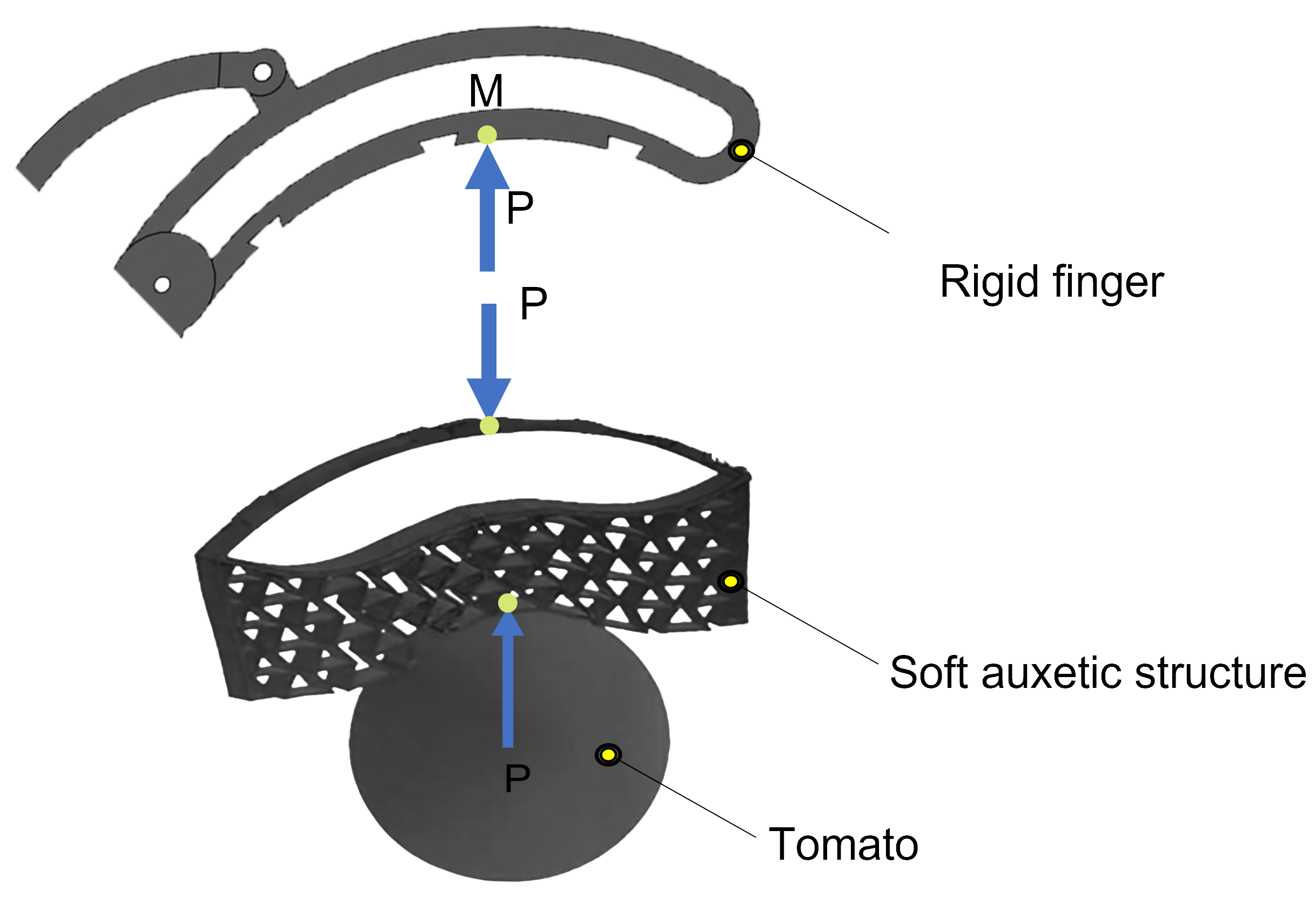}
       \caption{The figure shows the force of interaction of a single gripper finger as the gripper grasps the tomato with soft auxetic structure in contact during the picking operation for  static force analysis }
      \label{fig:static_force}
\end{figure}
\begin{figure}[htbp]
      \centering
      \includegraphics[width=0.7\linewidth]{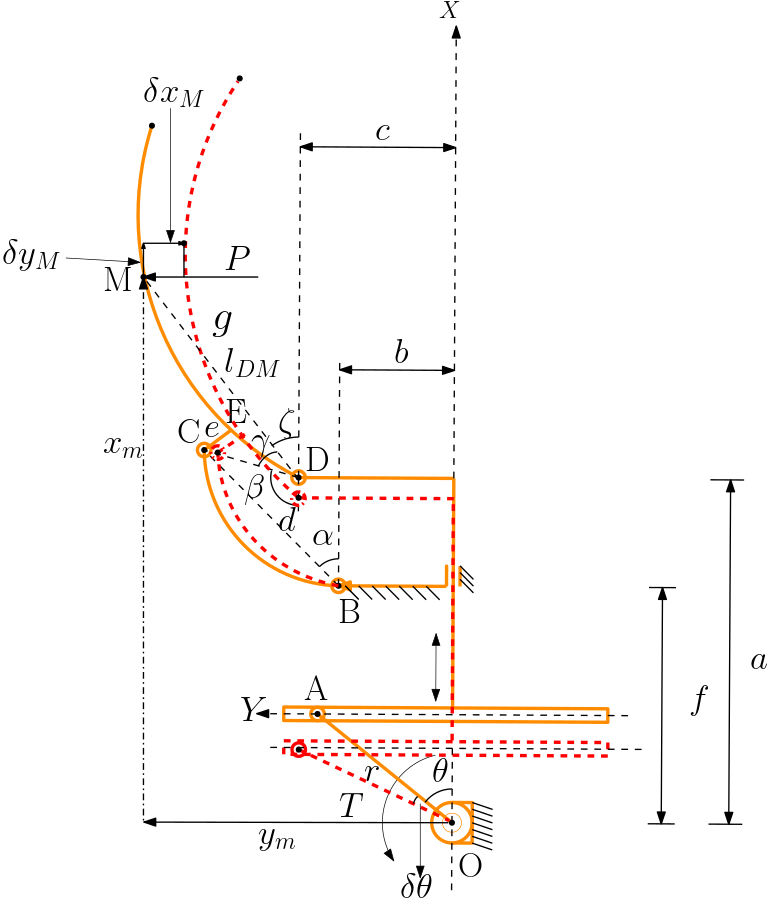}
       \caption{The figure shows the various kinematic parameters and the force acting on the rigid link of the finger, which is the reaction force coming from the auxetic structure support during the interaction with the tomato while grasping.}
      \label{fig:Gripper_kinematics}
\end{figure}
\begin{figure}[htbp]
      \centering
      \includegraphics[width=0.8\linewidth]{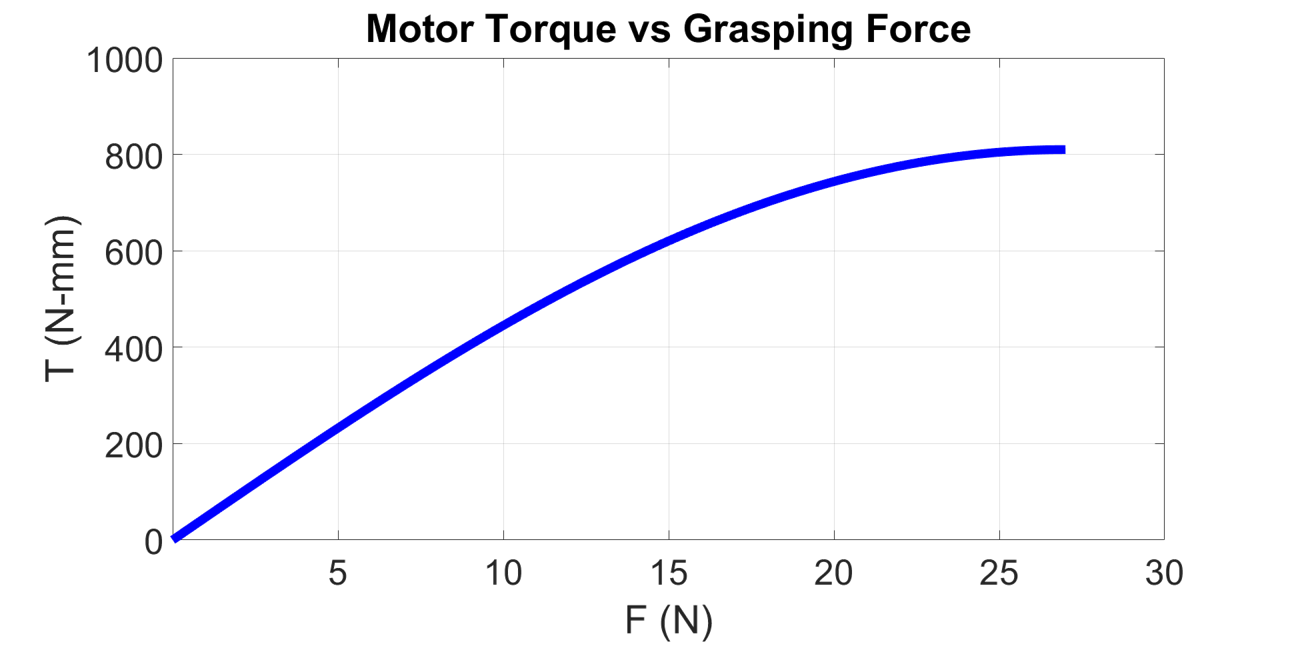}
       \caption{The figure shows the variation of motor torque required to provide the desired grasping force to the gripper fingers for holding the tomato }
      \label{fig:Force_torque_curve}
\end{figure}
\begin{figure*}[htbp]
      \centering
      \includegraphics[width=\linewidth]{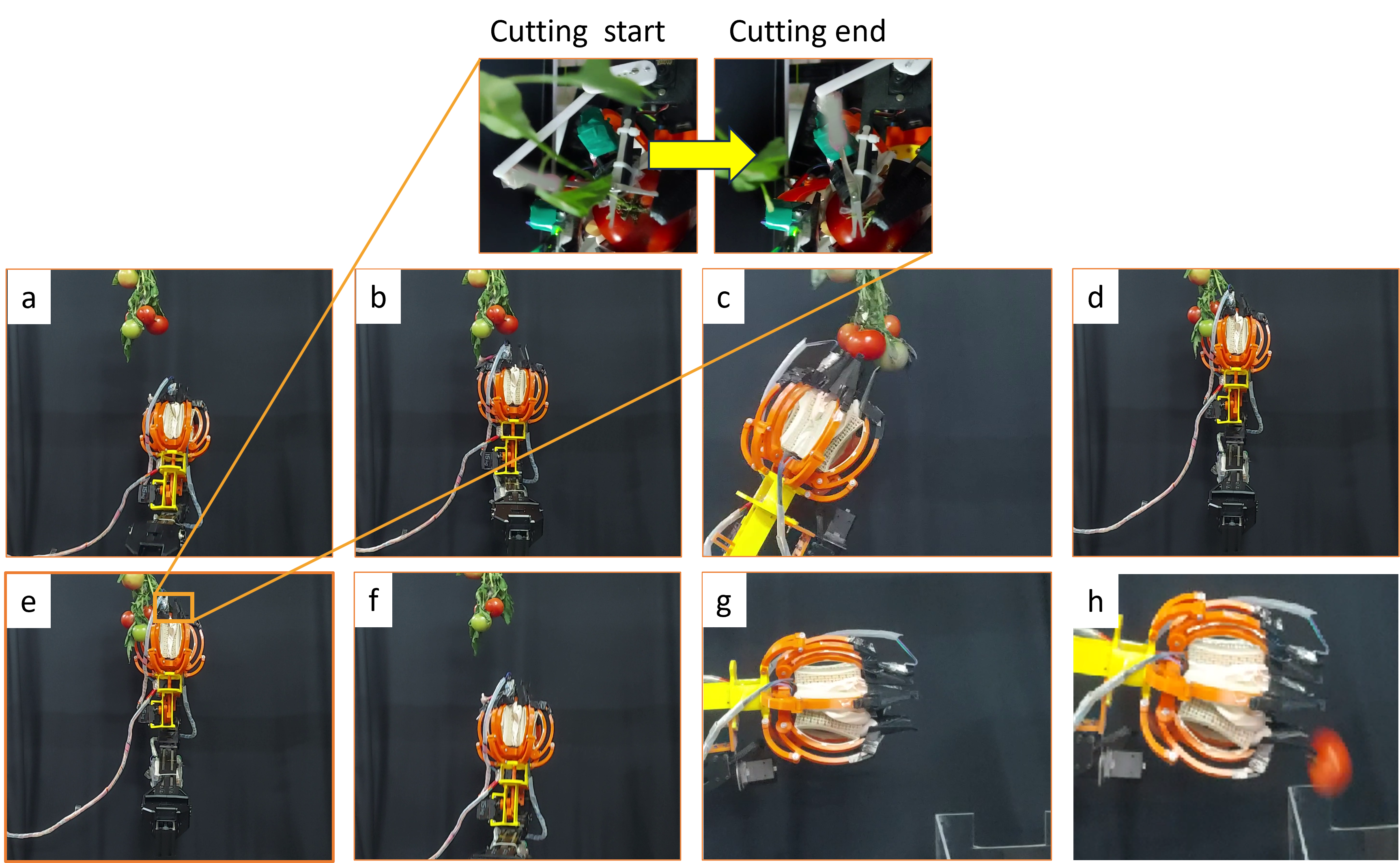}
       \caption{In this figure, the tomato picking operation is performed experimentally in the lab environment under controlled conditions.From (a-b): Gripper
Approached towards the tomato cluster (c-d): The target tomato is separated from the cluster (e). The cutting action of the tomato pedicel is performed (f)Soft grasping of the tomato under the influence of auxetic fingers (g)Gripper departs towards the punnet (h): Tomato  is released inside a punnet}
      \label{Picking_operation}
\end{figure*}
\begin{figure}[htbp]
      \centering
      \includegraphics[width=\linewidth]{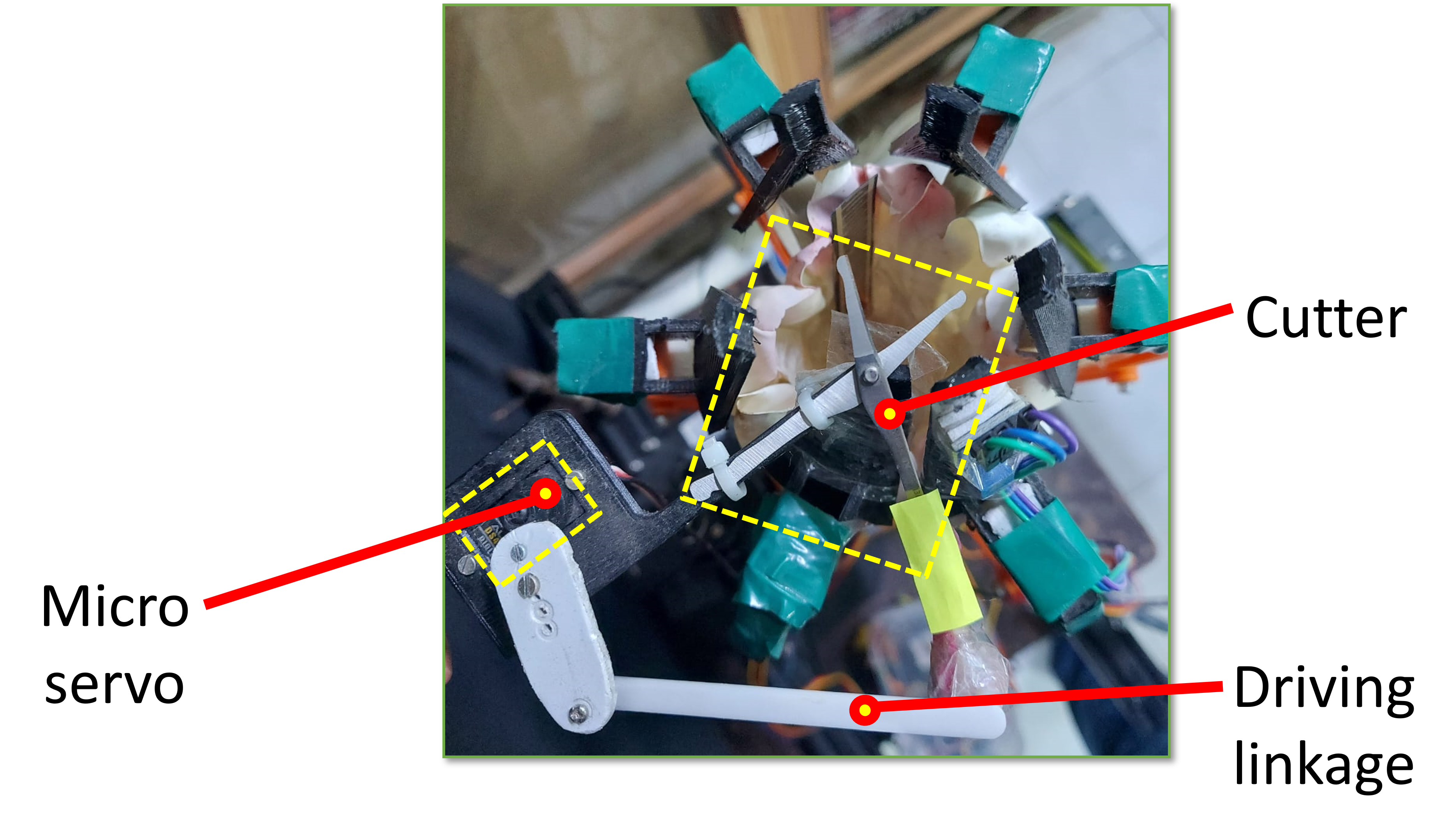}
       \caption{The figure shows the micro servo actuated cutter mechanism mounted over one of the separator leaves of the gripper, which will be actuated when the separator is in completely closed state }
      \label{fig:cutter_mechanism}
\end{figure}
\begin{figure}[htbp]
      \centering
      \includegraphics[width=\linewidth]{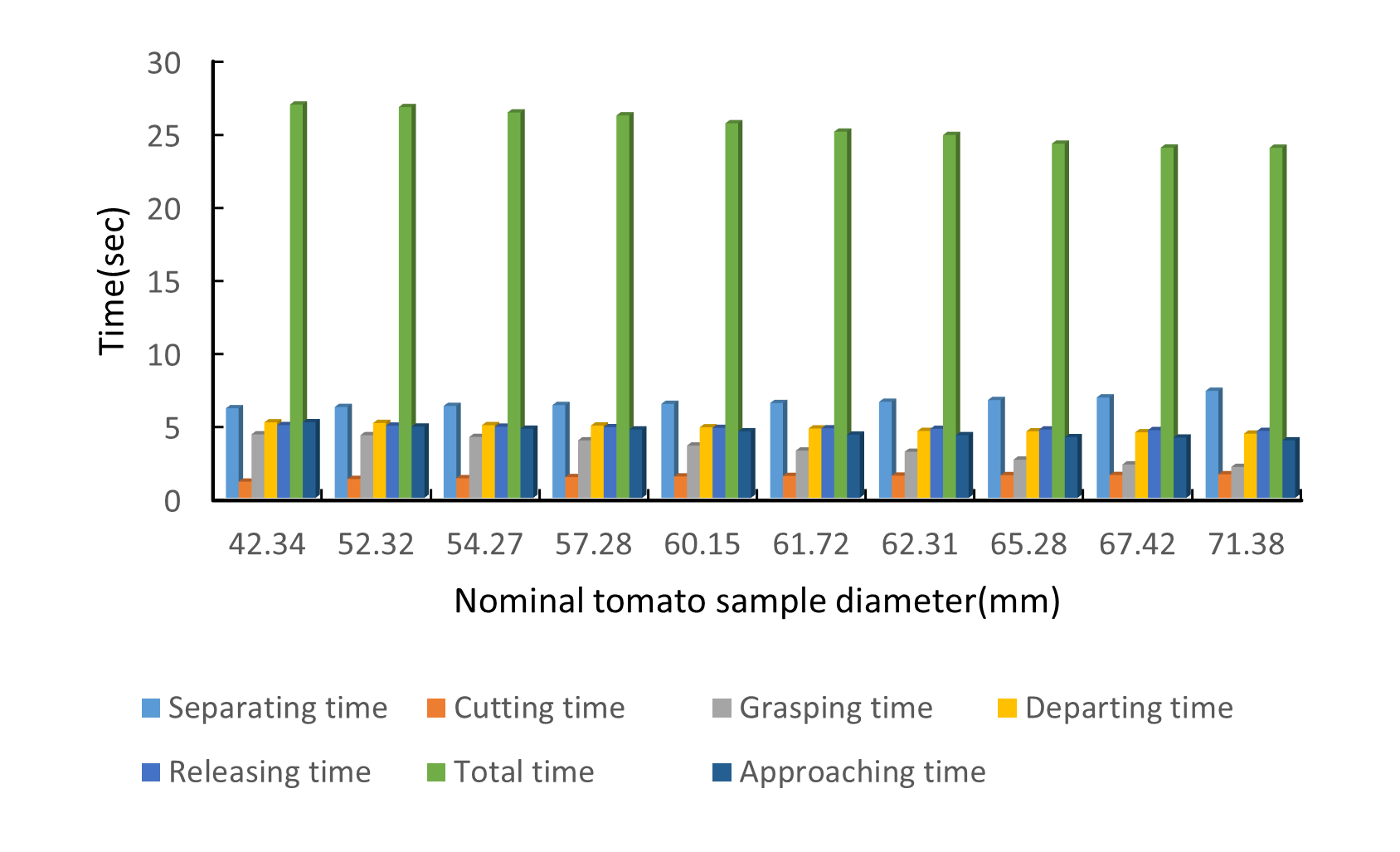}
       \caption{The figure shows the complete bar chart for the tomato picking operation, where for each tomato sample diameter, the corresponding set of the variation of approaching, separating, cutting, grasping, departing, releasing, and total time}
      \label{graph: Picking operation }
\end{figure}

\section{Tomato Semantic Segmentation with Tomato Center and Pedicel Keypoint Detection}
The study employs deep learning techniques for tomato semantic segmentation and keypoint detection. By distinguishing between ripe and unripe tomatoes, the segmentation process enables a robotic arm to determine which tomatoes are ready for harvest. The tomato's pedicel, or the point where the tomato attaches to the plant, is the focus of keypoint detection. This is essential for the robotic arm to precisely cut and pick the tomato without needlessly harming the plant or the fruit. We have also incorporated the tomato center with keypoint, which helps locate the center preciously for robotic arm manipulation. We have used Rob2Pheno dataset \cite{afonso21} and our own collected data in the field and laboratory. We have used Detectron-2 \cite{wu2019detectron2} for this purpose. Detectron-2 is a model built on Pytorch for segmentation and keypoint detection, is mentioned in the paper. The Adam optimizer trains this model at an initial learning rate of 0.001. The model is trained for 20,000 iterations on the larger dataset. The outcomes show that various experimental backbones can yield consistent outcomes across the dataset, with the ResNeXt-based model outperforming the ResNet-50-based model. The performance of different backbones is shown in table \ref{tab:result_detect2}. 

\begin{table*}[htbp]
\caption{Performance of Detectron-2 with different models }
\label{tab:result_detect2}
\resizebox{\textwidth}{!}{%
\begin{tabular}{|c|c|c|c|c|c|c|c|c|l|}
\hline
Backbone & Training Iterations & Learning Rate & Optimizer & Train Loss & Test Loss & mAP & Precision & Recall \\ \hline
ResNet-50 & 20,000 & 0.001 & Adam & 0.2 & 0.3 & 0.75 & 0.8 & 0.7 \\ \hline
ResNeXt & 20,000 & 0.001 & Adam & 0.1 & 0.2 & 0.8 & 0.85 & 0.75 \\ \hline
ResNet-101 & 20,000 & 0.001 & Adam & 0.15 & 0.25 & 0.77 & 0.82 & 0.72 \\ \hline
\end{tabular}%
}
\end{table*}
\section{Robotic Gripper Control Strategy for Tomato Handling: Workflow and Integration}
\begin{figure}[htbp]
    \centering
    \includegraphics[width=\linewidth]{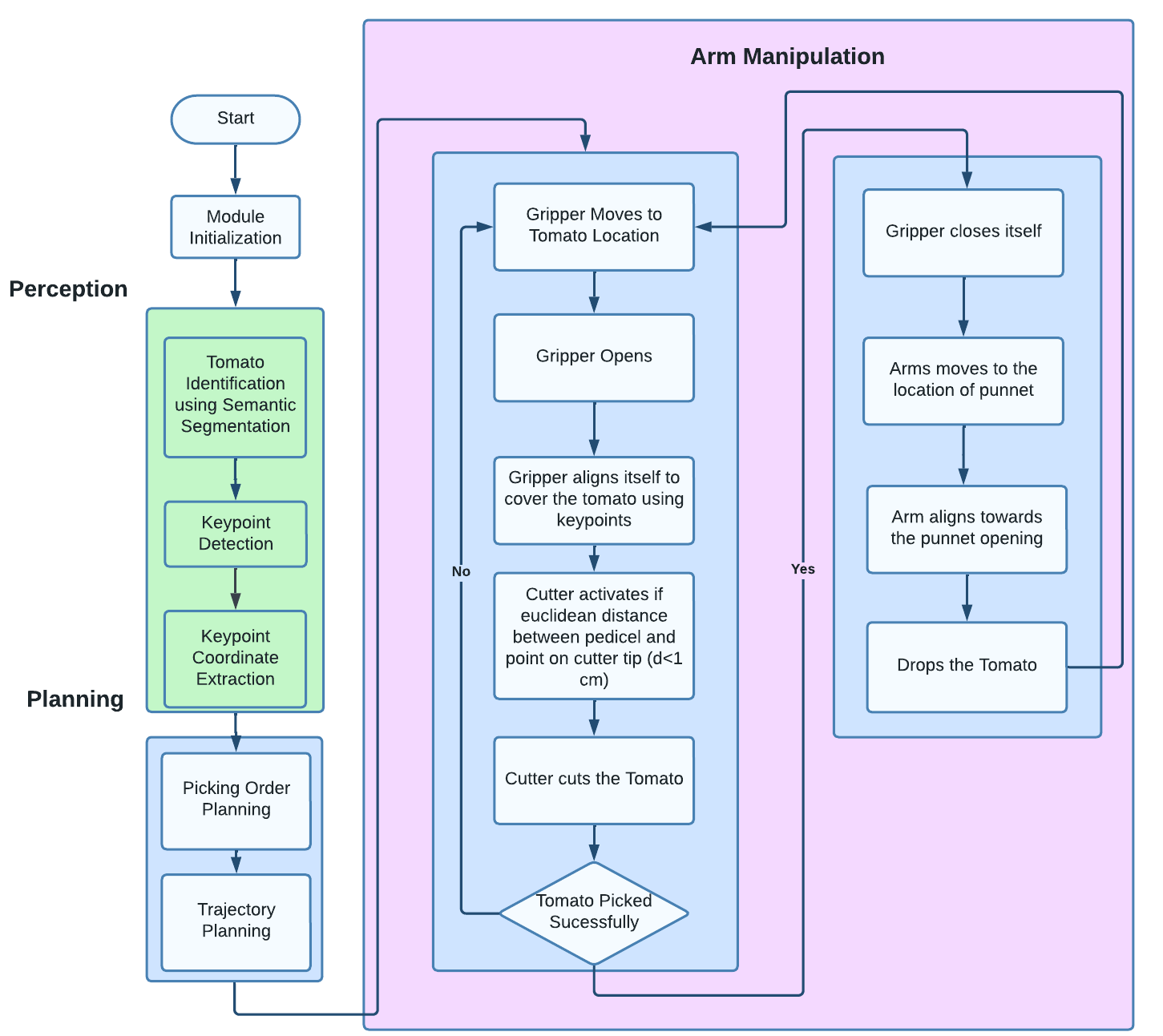}
    \caption{The figure shows the flow chart for the complete tomato picking operation using the proposed gripper on a robotic manipulator with different operation stages starting from perception, planning, arm manipulation and dropping in the punnet.}
    \label{fig:control_flow}
\end{figure}

Figure \ref{fig:control_flow} describes the gripper control strategy and trajectory. Trajectory planning is done through Particle  Swarm Optimization (PSO). PSO in robotic trajectory planning iteratively optimizes candidate paths. Particles, representing potential trajectories, adjust based on their best positions and the swarm's best, converging on collision-free and potentially efficient motions \cite{ekrem2023trajectory}.
Here is a concise explanation of each part of the chart:
\begin{itemize}
\item{\textbf{Machine Learning}}: This section focuses on the training of machine learning models, specifically those related to "Segmentation and Keypoint Models," utilizing the Detectron 2 framework. The models are subsequently validated in order to ascertain their accuracy and performance.
\item {\textbf{Integration:}} During the integration phase, data pertaining to RGB and Depth is acquired from the surrounding environment. The data is subsequently recorded in order to precisely synchronize the depth and color information. Following this, machine learning models are employed to make predictions on key points and carry out semantic segmentation in order to distinguish between ripe and unripe tomatoes. The outcomes are transferred to the subsequent stage known as the "2D to 3D Conversion" process.
\item {\textbf{Robotic Control:}} The process known as "2D to 3D Conversion" involves the transformation of 2D data, specifically image-based predictions, into a three-dimensional representation of tomatoes. The provided data is utilised in the context of "Path Planning," a process that involves the computation of an optimal trajectory for the robotic gripper. The calculated trajectory directs the "Control Systems" to effectively regulate the gripper's movements.
\item {I\textbf{mplementation:}} The implementation stage involves the practical application of control algorithms in a real-world environment. The initial step involves simulating and testing the system to verify its operational capabilities. Upon successful completion of this phase, the subsequent stage, referred to as "Real-world Implementation," is initiated. This phase entails the deployment of the robotic system for the practical handling of tomatoes.
\end{itemize}
\section{Experimental evaluation}
First, the camera system identifies the key point location in the target tomato crop through semantic segmentation. The first point is the center of the tomato to be recognized as the target tomato, and the second is at its pedicel, where the gripper's cutter has to make the final cut for detaching from the plant. Based upon this information obtained, the robotic arm with the gripper is given a command to move towards the desired location of the target tomato, after which the arm tends to push the target tomato, simultaneously opening the separator leaves to separate it from the nearby tomatoes in its vicinity constraining them to slide over the separator's conical frustum surface while the tomato to be picked moves inside the gripper. As the tomato reaches inside the gripper up to a particular depth level, a specified distance between the key point on the cutter and the pedicel is calculated, and based upon the information of the value of that distance if reached, then the cutter will be actuated by the micro servo for the successful pedicel cutting operation. After this step, the tomato will fall inside the gripping area surrounded by a flexible latex basket backed by six auxetic structure-based fingers for performing the grasping operation. After the grasping process, the robotic manipulator will move towards the punnet to drop it inside successfully. All these operations together add up to make a complete one-picking operation cycle. In Fig. \ref{Picking_operation}, it can be observed how the tomato picking operation is conducted using a robotic gripper mounted over the arm. In this figure, the various steps of the tomato-picking process are shown. Here, the cutting action image in the inset shown is taken separately rather than simultaneously to show how the cutting action takes place because a multi-camera system is not used to show the process from various angles. The micro servo actuated cutter mechanism used to cut the tomato pedicel can be seen in figure \ref{fig:cutter_mechanism}. The tomato picking operation can be seen in figure \ref{graph: Picking operation } shows the variation of various operation times during the tomato picking operations. Also, it was found that the average success rate of tomato picking during the ten trials of the experiment was approximately 80\% for every trial, the success rate is calculated, which can be seen in table \ref{tab:result_detect2}.
\begin{figure}[htbp]
    \centering
    \includegraphics[width=0.8\columnwidth]{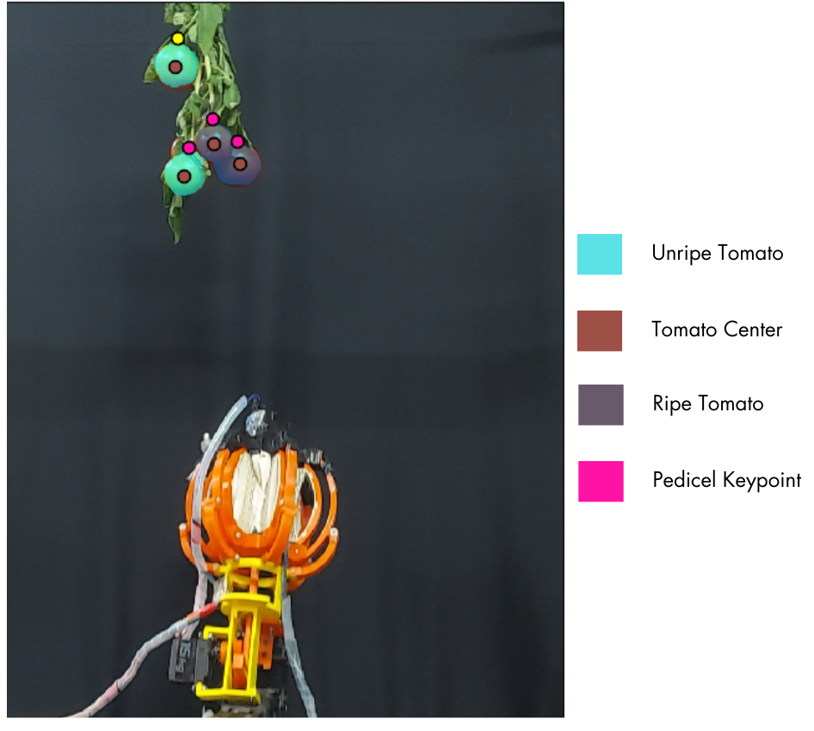}
    \caption{This figure depicts the semantic segmentation of tomato fruit, which includes segmenting ripe and unripe tomatoes as well as detecting the keypoint of the pedicel and tomato centre.}
    \label{fig:enter-label}
\end{figure}
\section{CONCLUSIONS}
The proposed work described an approach towards developing the tomato harvesting robotic system, which can harvest tomatoes mainly in a cluttered environment of leaves, tomatoes, and plant branches. Here, the harvesting process is demonstrated on the cluster of tomatoes to separate the target tomato from it. The tomato picking process, which involves various operations, such as separating the target tomato from the cluster, cutting its pedicel, grasping, holding it in place, and dropping it to the desired location, is performed during the single operation cycle. The primary contribution of this work is the design and development of a novel hybrid robotic gripper with a soft grasping auxetic structure and a rigid exoskeleton for good gripping strength. Additionally, a vision system based on a depth camera implements the fruit identification process. It incorporates keypoint detection of the tomato's pedicel and tomato for localization in a variable ambient light environment and deep learning-based semantic segmentation of ripe and unripe tomatoes. In addition, robust trajectory planning of the robotic arm based on input from the vision system and control of robotic gripper movements are carried out for secure tomato handling. The variation of various operation times w.r.t. the nominal tomato diameter for a single tomato picking cycle is presented. These steps include approaching, separating, cutting, grasping, departing, and releasing, and it was found that all these operations collectively make one complete operation cycle, which takes an average time of approximately 24.34 seconds. The experimental results of the tomato-picking process are presented, and it was found that the average success rate of the tomato-picking operation is approximately 80\%. However, our system could be better in terms of performing cutting operations, as sometimes it was found that during the pedicel cutting phase, the pedicel missed the cutting zone of the cutter, so the tomato picking operation failed. Future work will involve redesigning the pedicel-cutting mechanism and optimizing the gripper's design.

\section*{ACKNOWLEDGMENT}
This research was supported by the Department of Science and Technology, Government of India, under project number DST/ME/2020009.

\addtolength{\textheight}{-0cm} 

\bibliographystyle{ieeetr}
\bibliography{references.bib}
\clearpage
\end{document}